\documentclass[11pt,a4paper]{article}
\usepackage[hyperref]{eacl2021}
\usepackage{times}
\usepackage{latexsym}

\usepackage{microtype}

\usepackage{url}
\usepackage{amsfonts, amssymb}       
\usepackage{nicefrac}       
\usepackage{microtype}      
\usepackage{graphicx}
\usepackage{capt-of}
\usepackage{float}
\usepackage{macros}
\usepackage{mathtools}
\usepackage{array}
\usepackage{url}
\usepackage{multirow}
\usepackage{wrapfig}
\usepackage{amsmath}

\aclfinalcopy 

\newcommand\proformer{ProFormer }

\title{ProFormer: Towards On-Device LSH Projection Based Transformers}

\author{Chinnadhurai Sankar \\
  Mila, Universit\'e de Montr\'eal\Thanks{Work done during internship at Google}\\
  Montreal, QC, Canada\\
  \texttt{chinnadhurai@gmail.com} \\\And
 
  Sujith Ravi  \\
  Amazon Alexa\Thanks{Work done while at Google AI} \\
  Sunnyvale, CA, USA  \\
  \texttt{sravi@sravi.org}\\ \And
  
  Zornitsa Kozareva  \\
  Google \\
   Mountain View, CA, USA \\
  \texttt{zornitsa@kozareva.com} \\}
  
\date{}

\begin{document}
\maketitle
\begin{abstract}
At the heart of text based neural models lay word representations, which are powerful but occupy a lot of memory making it challenging to deploy to devices with memory constraints such as mobile phones, watches and IoT. To surmount these challenges, we introduce \proformer  -- a projection based transformer architecture that is faster and lighter making it suitable to deploy to memory constraint devices and preserve user privacy. We use LSH projection layer to dynamically generate word representations on-the-fly without embedding lookup tables leading to significant memory footprint reduction from  $\mathcal{O}(V.d)$ to $\mathcal{O}(T)$, where $V$ is the vocabulary size, $d$ is the embedding dimension size and $T$ is the dimension of the LSH projection representation. We also propose a \textit{local projection attention (LPA)} layer, which uses self-attention to transform the input sequence of $N$ LSH word projections into a sequence of $N/K$ representations reducing the computations quadratically by $\mathcal{O}(K^2)$. 

We evaluate ProFormer on multiple text classification tasks and observed improvements over prior state-of-the-art on-device approaches for short text classification and comparable performance for long text classification tasks. ProFormer is also competitive with other popular but highly resource-intensive approaches like BERT and even outperforms small-sized BERT variants with significant resource savings -- reduces the embedding memory footprint from 92.16 MB to 1.7 KB and requires $16 \times$ less computation overhead, which is very impressive making it the fastest and smallest on-device model. 

\end{abstract}

\section{Introduction}
Transformers \citep{vaswani2017attention} based architectures like BERT \citep{bert2018}, XL-net \citep{xlnet}, GPT-2 \citep{gpt2}, MT-DNN \citep{mtdnn}, RoBERTA \citep{roberta} reached state-of-the-art performance on tasks like machine translation \citep{bert_nmt}, language modelling \citep{gpt2}, text classification benchmarks like GLUE \citep{glue}. However, these models require huge amount of memory and need high computational requirements making it hard to deploy to small memory constraint devices such as mobile phones, watches and IoT. Recently, there have been interests in making BERT lighter and faster \citep{distill_bert, prune_bert}. In parallel, recent on-device works like SGNN \cite{sgnn}, SGNN++ \cite{ravi-kozareva-2019-device} and \cite{sankar-etal-2019-transferable} produce lightweight models with extremely low memory footprint. They employ a modified form of LSH projection to dynamically generate a fixed binary projection representation, $\mathbb{P}(x) \in [0,1]^{T}$ for the input text $x$ using word or character n-grams and skip-grams features, and a 2-layer MLP $+$ softmax layer for classification. As shown in \cite{sgnn} these models are suitable for short sentence lengths as they compute $T$ bit LSH projection vector to represent the entire sentence. However, \cite{proseqo2019} showed that such models cannot handle long text due to significant information loss in the projection operation.

On another side, recurrent architectures represent long sentences well, but the sequential nature of the computations increases latency requirements and makes it difficult to launch on-device. Recently, self-attention based architectures like BERT \citep{bert2018} have demonstrated remarkable success in capturing long term dependencies in the input text via purely attention mechanisms. BERT’s model architecture is a multi-layer bidirectional Transformer encoder based on the original implementation in \citep{vaswani2017attention}. The self-attention scores can be computed in parallel as they do not have recurrent mechanisms. But usually these architectures are very deep and the amount of computation is quadratic in the order of $\mathcal{O}(L \cdot N^2)$, where $L$ is the number of layers (Transformer blocks) and $N$ is the input sentence length. Straightforward solutions like reducing the number of layers is insufficient to launch transformers on-device due to the large memory and quadratic computation requirements.

In this paper, we introduce a projection-based neural architecture \proformer that is designed to (a) be efficient and learn compact neural representations (b) handle out of vocabulary words and misspellings (c) drastically reduce embedding memory footprint from hundreds of megabytes to few kilobytes and (d) reduce the computation overhead \textit{quadratically} by introducing a local attention layer which reduces the intermediate sequence length by a constant factor, $K$. We achieve this by bringing the best of both worlds by combining LSH projection based representations (for low memory footprint) and self-attention based architectures (to model dependencies in long sentences). To tackle computation overheard in the transformer based models, we reduce the number of self-attention layers and additionally introduce an intermediate local projection attention (LPA) to quadratically  reduce the number of self-attention operations. The main contributions of our paper are: \vspace{-0.2cm}

\begin{itemize} 
\item We propose novel on-device neural network called \proformer which combines LSH projection based text representations, with transformer architecture and locally projected self-attention mechanism that captures long range sentence dependencies while yielding low memory footprint and low computation overhead. \vspace{-0.2cm}

\item \proformer reduces the computation overhead $\mathcal{O}(L \cdot N^2)$ and latency in multiple ways: by reducing the number of layers $L$ from twelve to two and introducing new local projection attention layer that decreases number of self-attention operations by a quadratic factor. \vspace{-0.2cm}

\item \proformer is light weigh compact on-device model, while BERT on-device still needs huge embedding table ( $92.16$ MB for $V=30k$, $d=768$) with number of computation flops in the order of $\mathcal{O}(L \cdot N^2)$, where $L$ is the number of layers, $N$ is the number of words in the input sentence.  \vspace{-0.2cm}

\item We conduct empirical evaluations and comparisons against state-of-the-art on-device and prior deep learning approaches for short and long text classification. Our model \proformer reached state-of-art performance for short text and comparable performance for long texts, while maintaining small memory footprint and computation requirements.

\end{itemize}

\section{ProFormer: LSH Projection based Transformers}
\begin{figure}[!htbp]
\centering
\includegraphics[scale=0.55]{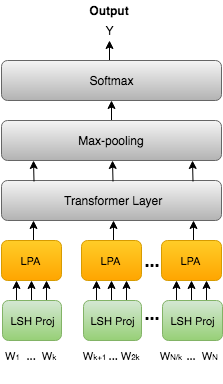}
\caption{ProFormer: Our \textbf{Pro}jection Trans\textbf{former} Network Architecture}
\label{fig:ProFormer}
\end{figure}

In this section, we show the overall architecture of \proformer in Figure \ref{fig:ProFormer}. \proformer consists of multiple parts: (1) word-level Locality Sensitive Hashing (LSH) projection layer, (2) local projection attention (LPA) layer, (3) transformer layer \citep{bert2018} and (4) a max-pooling $+$ classifier layer. Next, we describe each layer in detail.

\subsection{LSH Projection Layer}
It is a common practice to represent each word in the input sentence, $\mathbf{x} = [w_1, w_2, \, \cdots \,, w_N]$ as an embedding vector based on its one-hot representation. Instead, we adopt LSH projection layer from \citep{DBLP:journals/corr/abs-1708-00630, pmlr-v97-ravi19a} which dynamically generates a $T$ bit representation, $\mathbb{P}(w_i) \in [0,1]^T$ for the input word, $w_i$ based on its morphological features like n-grams, skip-grams from the current and context words, parts-of-speech tags, etc. 

Since the LSH projection based approach does not rely on  embedding lookup tables to compute word representation, we obtain significant memory savings of the order, $O(V \cdot d)$, where $V$ is the vocabulary size and $d$ is the embedding dimension. For instance, the embedding look-up table occupies 92.16 MB ($V=30k$, $d=768$ \citep{bert2018}), while the LSH projection layer requires only $\approx$ 1.7 KB ($T=420$) as shown in Table \ref{table:bert_vs_ProFormer_comp}.

\begin{table}[!htbp]
\scalebox{0.7}{
\begin{tabular}{lcc}
\hline
\textbf{Models} & \textbf{Embedding memory} & \textbf{Computations} \\
\hline
BERT & $\mathcal{O}(V.d)$  &  $\mathcal{O}(N^2)$\\
\proformer (our model) & $\mathcal{O}(T)$ & $\mathcal{O}(N^2/K^2)$ \\
\hline
\end{tabular}
}
\caption{Memory and computations overhead comparison between BERT \citep{bert2018} and ProFormer (our model). $N$ is the number of words in the input. For $V=30k,\,d=768,\,T=420$, BERT's embedding table occupies $92.16$ MB while \proformer requires only $\mathbf{1.7}$ KB. For $K=4$, we reduce the BERT computation overhead by $\mathbf{16}$ times.} 
\label{table:bert_vs_ProFormer_comp}
\end{table}

\subsection{Local Projection Attention (LPA) Layer} 

\begin{figure}[!htbp]
\centering
\includegraphics[scale=0.55]{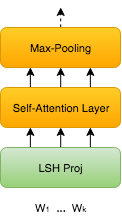}
\caption{Local Projection Attention (LPA) layer.}
\label{fig:LPA}
\end{figure}

The LPA layer shown in Figure \ref{fig:LPA} consists of a single layer multi-headed self-attention layer similar to the Transformer architecture in \citep{vaswani2017attention} followed by a max-pooling layer yielding a compressed representation of $K$ input words, $[w_1,\,w_2,\, \cdots w_K]$. 

\noindent The LPA layer transforms the $N$ word-level projections, $\mathbb{P}(w_{i})$ to a sequence of $N/K$ representations as in Equation \ref{eq:lpa}.
\vspace{-0.1cm}
\begin{multline}
\label{eq:lpa}
\scriptstyle [\mathbb{P}(w_1),\, \cdots \, \mathbb{P}(w_N)]_{N} \,\, \longrightarrow \\ \scriptstyle [\mathcal{LPA}(\mathbb{P}(w_{1:K})),\, \cdots \, \mathcal{LPA}(\mathbb{P}(w_{N/K:N}))
]_{N/K}   
\end{multline}
\noindent where $\mathcal{LPA}$ consists of the self-attention and max-pooling operation, $K$ is a \textit{Group factor}\footnote{We choose $K$ such that $N$ is divisible by $K$.}. We equally divide the $N$ word-level LSH projection representations into $N/K$ groups of size $K$. The LPA layer compresses each group of $K$ word representations into $\mathcal{LPA}(\mathbb{P}(w_{1:K})) \in \mathbb{R}^d$ yielding $N/K$ representations in total. The LPA layer reduces the self-attention computation overhead in the subsequent transformer layer \citep{vaswani2017attention} by $\mathcal{O}(K^2)$.

\subsection{Transformer Layer}
This layer consists of 2-layer bidirectional Transformer encoder based on the original implementation described in \citep{vaswani2017attention}.
This layer transforms the $N/K$ input representations from the LPA layer described in the previous sub-section into $N/K$ output representations. In this layer, we reduce both the computation overhead and memory footprint by reducing the number of layers from $L$ to $2$ reducing the computation overhead by $\mathcal{O}(L/2)$ ($6$ times in the case of 12-layer \textit{BERT-base} model).

\subsection{Max-Pooling and Classification Layer}
We summarize the $N/K$ representations from the transformer layer to get a single $d$ dimensional vector by max-pooling across the $N/K$ time-steps, followed by a softmax layer to predict the output class $Y$. 

\section{Datasets \& Experimental Setup}
In this section, we describe our datasets and experimental setup. We use text classification datasets from state-of-the-art on-device evaluations such as: MRDA \citep{MRDA} and ATIS \citep{atis}, AG News \cite{zhang2015text} and Yahoo! Answers \cite{zhang2015text}. Table \ref{table:datasets} shows the characteristics of each dataset.


\begin{table}[!htbp]
\scalebox{0.60}{
\begin{tabular}{lcccc}
\hline
Tasks & \# Classes & Avg-len & Train & Test\\
\hline
MRDA (Dialog act) & $6$ & $8$ & 78k & 15k\\
ATIS (Intent prediction) & $21$ & $11$ & $4.4k$ & $0.89k$ \\

AG (News Categorization) & 4 & 38 & 120k & 7.6k \\
Y!A (Yahoo! Answers Categorization) & 10 & 108 & 1400k & 60k \\
\hline
\end{tabular}
}
\caption{Classification Dataset Characteristics}
\label{table:datasets}
\end{table}

\noindent We train \proformer on multiple classification tasks individually and report \textit{Accuracy} on corresponding test sets.  
We fix the projection size, $T=420$,
n-gram size=5, skip-gram size=1 for the LSH projection operation, $\mathbb{P}$. For the LPA layer, We experiment with two values for $K=1, 4$, where $K=1$ corresponds to the null operation in the LPA layer which just passes the word LSH projection representation to the Transformer layer. For the transformer layer, we fix the number of layers, $L=2$ and set all layer sizes, $d=768$ (including the intermediate size for the dense layer).\footnote{The rest of the parameters are same as the one used in $bert\_config.json$ in \textit{BERT-base} model \citep{bert2018}}  

We compare our model with previous state of the art neural architectures, including on-device approaches. We also fine-tune the pretrained 12-layer \textit{BERT-base} model \citep{bert2018} on all classification tasks and compare to our model. \textit{BERT-base} consists 12-layers of transformer blocks \citep{vaswani2017attention} and is pretrained in an unsupervised manner on a large corpus (BooksCorpus \citep{bookcorpus} and English WikiPedia) using masked-language model objective. 
We fine-tune the pretrained  \textit{BERT-base} \citep{bert2018} to each of the classification tasks. For training, we use Adam with learning rate of $1$e-$4$, $\beta_1$=$0.9$,  $\beta_2$=$0.999$, $L2$ weight decay of $0.01$, learning rate warmup over the first $10,000$ steps, and linear decay of the learning rate. We use dropout probability of $0.1$ on all layers and training batch size of $256$. For further comparison, we also trained much smaller BERT baselines with 2-layers of transformer blocks and smaller input embedding sizes.

\section{Results}

Tables \ref{table:short_text_classification} and \ref{table:long_text_classification} show the results on the ATIS \& MRDA short text classification and AG \& Y!A long text classification tasks. We compare our approach, \proformer against prior state-of-the-art on-device works, fine-tuned \textit{BERT-base}, smaller 2-layer \textit{BERT} variants and other non-on-device neural approaches. 

Overall, our model \proformer improved upon non-on-device neural models while keeping very small memory footprint and high accuracy. This is very impressive since \proformer can be directly deployed to memory constraint devices like phones, watches and IoT while still maintaining high accuracy. \proformer also improved upon prior on-device state-of-the-art neural approaches like SGNN \cite{sgnn} and SGNN++ \cite{ravi-kozareva-2019-device} reaching over 35\% improvement on long text classification. Similarly it improved over on-device ProSeqo \cite{proseqo2019} models for all datasets and reached comparable performance on MRDA. In addition to the quality improvements, \proformer also keeps  smaller memory footprint than ProSeqo, SGNN and SGNN++.

In addition to the non-on-device and on-device neural comparisons, we also compare against \textit{BERT-base} and other smaller variants. Our experiments show that \proformer outperforms the small \textit{BERT} baselines on all tasks. Moreover, although the 12-layer fine-tuned \textit{BERT-base} \citep{bert2018} model converged to the state-of-the-art in almost all of the tasks, \proformer converges to $ \approx 97.2$\% \textit{BERT-base}'s performance on an average while occupying only $13$\% of \textit{BERT-base's} memory. \proformer has $14.4$ million parameters, while BERT-base has $110$ million. For fair comparison, we also test \proformer with $K=4$, which only occupies $38.4$\% the memory footprint of $2$-layer BERT-base model and reduces the computation overhead by $16$ times. The embedding look up table occupies nearly $23$ million parameters out of $38$ million parameters in the $2$-layer BERT model. We notice that $K$=$4$ model performs slightly worse than $K$=$1$ indicating information loss in the LPA layer. Overall, our experiments demonstrate that \proformer reaches better performances that prior non-on-device and on-device neural approaches, and comparable performance to BERT-base models while preserving smaller memory footprint.

\begin{table}[!htbp]
\centering
\scalebox{0.7}{
\begin{tabular}{lcc}
\hline
Models & MRDA & ATIS\\
\hline
\textbf{\proformer ($K$=1) (our model)} & \textbf{89.3} & \textbf{98.2}\\
\textbf{\proformer ($K$=4) (our model)} & 86.7 & 97.0 \\

BERT-base + fine-tuned \small{\citep{bert2018}} & \textbf{90.1} & \textbf{98.3} \\
\hspace{25pt} {\it (12-layers, embedding size = 768)} & & \\

BERT {\it (2-layer, embedding size = 560)} & 77.0 & 94.0 \\
BERT {\it (2-layer, embedding size = 840)} & 76.8 & 95.0 \\

ProSeqo \small{\cite{proseqo2019}}(on-device) & \textbf{90.1}    &  97.8    \\
SGNN++ \small{\cite{ravi-kozareva-2019-device}}(on-device)& 87.3 & 93.7 \\
SGNN \small{\cite{sgnn}}(on-device) &  86.7 & 88.9   \\

\hline 
RNN\small{\cite{khanpour-guntakandla-nielsen:2016:COLING}} & 86.8 &  -  \\

RNN+Attention\small{\cite{ortega-vu:2017:W17-55}} &  84.3  & -   \\
\small{CNN\small{\cite{lee-dernoncourt:2016:N16-1}}} &84.6 & -  \\

GatedIntentAtten.\small{\cite{N18-2118}}  &    - & 94.1   \\

GatedFullAtten.\small{\cite{N18-2118}}  &    - & 93.6   \\

JointBiLSTM\small{\cite{dilek}}  &  - & 92.6   \\

Atten.RNN\small{\cite{liu}}  &  - &91.1   \\
\hline
\end{tabular}
}
\caption{Short text classification results.}
\label{table:short_text_classification}
\end{table}
\vspace{-0.2cm}

\begin{table}[!htbp]
\centering
\scalebox{0.7}{
\begin{tabular}{lcc}
\hline
Models & AG & Y!A\\
\hline
\textbf{\proformer ($K$=1) (our model)} & \textbf{92.0} & \textbf{72.8}\\
\textbf{\proformer ($K$=4) (our model)} & 91.5 & 71.1 \\

BERT-base + fine-tuned \small{\citep{bert2018}} & \textbf{94.5} & \textbf{73.8} \\
\hspace{25pt} {\it (12-layers, embedding size = 768)} & & \\

BERT {\it (2-layer, embedding size = 560)} & 82.3 & - \\
BERT {\it (2-layer, embedding size = 840)} & 83.3 & - \\

ProSeqo \small{\cite{proseqo2019}}(on-device)  &  91.5 &    72.4\\
SGNN \small{\cite{sgnn}(on-device})  & 57.6  & 36.5  \\
\hline
FastText-full \cite{fasttext}&  \textbf{92.5}  &  72.3  \\
CharCNNLargeWithThesau.\small{\cite{Zhang:2015}} & 90.6  & 71.2\\
CNN+NGM \cite{Bui2018} & 86.9 & -   \\
LSTM-full \cite{Zhang:2015} &86.1  &  70.8  \\

\hline
\end{tabular}
}
\caption{Long text classification results.}
\label{table:long_text_classification}
\end{table}
\vspace{-0.1cm}

\section{Conclusion}
We proposed a novel on-device neural network ProFormer, which combines LSH projection based text representations, with trans-former architecture and locally projected self-attention mechanism that captures long range sentence  dependencies. Overall, \proformer  yields low memory footprint and reduces computations quadratically. In series of experimental evaluations on short and long text classifications we show that \proformer improved upon prior neural models and on-device work like SGNN \cite{sgnn}, SGNN++ \cite{ravi-kozareva-2019-device} and ProSeqo \cite{proseqo2019}. \proformer reached comparable performance to our BERT-base implementation, however it produced magnitudes more compact models than BERT-base. This is very impressive showing both effectiveness and compactness of our neural model. 

\bibliography{anthology,eacl2021}

\begin{thebibliography}{29}
\expandafter\ifx\csname natexlab\endcsname\relax\def\natexlab#1{#1}\fi

\bibitem[{Arivazhagan et~al.(2019)Arivazhagan, Bapna, Firat, Lepikhin, Johnson,
  Krikun, Chen, Cao, Foster, Cherry, Macherey, Chen, and Wu}]{bert_nmt}
Naveen Arivazhagan, Ankur Bapna, Orhan Firat, Dmitry Lepikhin, Melvin Johnson,
  Maxim Krikun, Mia~Xu Chen, Yuan Cao, George Foster, Colin Cherry, Wolfgang
  Macherey, Zhifeng Chen, and Yonghui Wu. 2019.
\newblock \href {http://arxiv.org/abs/1907.05019} {Massively multilingual
  neural machine translation in the wild: Findings and challenges}.
\newblock \emph{CoRR}, abs/1907.05019.

\bibitem[{Bui et~al.(2018)Bui, Ravi, and Ramavajjala}]{Bui2018}
Thang~D. Bui, Sujith Ravi, and Vivek Ramavajjala. 2018.
\newblock Neural graph learning: Training neural networks using graphs.
\newblock In \emph{Proceedings of the Eleventh ACM International Conference on
  Web Search and Data Mining}, WSDM '18, pages 64--71.

\bibitem[{Devlin et~al.(2018)Devlin, Chang, Lee, and Toutanova}]{bert2018}
Jacob Devlin, Ming{-}Wei Chang, Kenton Lee, and Kristina Toutanova. 2018.
\newblock \href {http://arxiv.org/abs/1810.04805} {{BERT:} pre-training of deep
  bidirectional transformers for language understanding}.
\newblock \emph{CoRR}, abs/1810.04805.

\bibitem[{Goo et~al.(2018)Goo, Gao, Hsu, Huo, Chen, Hsu, and Chen}]{N18-2118}
Chih-Wen Goo, Guang Gao, Yun-Kai Hsu, Chih-Li Huo, Tsung-Chieh Chen, Keng-Wei
  Hsu, and Yun-Nung Chen. 2018.
\newblock Slot-gated modeling for joint slot filling and intent prediction.
\newblock In \emph{Proceedings of the 2018 Conference of the North American
  Chapter of the Association for Computational Linguistics: Human Language
  Technologies, Volume 2 (Short Papers)}, pages 753--757. Association for
  Computational Linguistics.

\bibitem[{Hakkani-Tur et~al.(2016)Hakkani-Tur, Tur, Celikyilmaz, Chen, Gao,
  Deng, and Wang}]{dilek}
Dilek Hakkani-Tur, Gokhan Tur, Asli Celikyilmaz, Yun-Nung~Vivian Chen, Jianfeng
  Gao, Li~Deng, and Ye-Yi Wang. 2016.
\newblock Multi-domain joint semantic frame parsing using bi-directional
  rnn-lstm.
\newblock In \emph{Proceedings of The 17th Annual Meeting of the International
  Speech Communication Association (INTERSPEECH 2016)}.

\bibitem[{Joulin et~al.(2016)Joulin, Grave, Bojanowski, Douze, J{\'{e}}gou, and
  Mikolov}]{fasttext}
Armand Joulin, Edouard Grave, Piotr Bojanowski, Matthijs Douze, Herv{\'{e}}
  J{\'{e}}gou, and Tomas Mikolov. 2016.
\newblock \href {http://arxiv.org/abs/1612.03651} {Fasttext.zip: Compressing
  text classification models}.
\newblock \emph{CoRR}, abs/1612.03651.

\bibitem[{Khanpour et~al.(2016)Khanpour, Guntakandla, and
  Nielsen}]{khanpour-guntakandla-nielsen:2016:COLING}
Hamed Khanpour, Nishitha Guntakandla, and Rodney Nielsen. 2016.
\newblock Dialogue act classification in domain-independent conversations using
  a deep recurrent neural network.
\newblock In \emph{Proceedings of COLING 2016, the 26th International
  Conference on Computational Linguistics: Technical Papers}, pages 2012--2021.

\bibitem[{Kozareva and Ravi(2019)}]{proseqo2019}
Zornitsa Kozareva and Sujith Ravi. 2019.
\newblock \href {https://doi.org/10.18653/v1/D19-1402} {{P}ro{S}eqo: Projection
  sequence networks for on-device text classification}.
\newblock In \emph{Proceedings of the 2019 Conference on Empirical Methods in
  Natural Language Processing and the 9th International Joint Conference on
  Natural Language Processing (EMNLP-IJCNLP)}, pages 3892--3901.

\bibitem[{Lee and Dernoncourt(2016)}]{lee-dernoncourt:2016:N16-1}
Ji~Young Lee and Franck Dernoncourt. 2016.
\newblock Sequential short-text classification with recurrent and convolutional
  neural networks.
\newblock In \emph{Proceedings of the 2016 Conference of the North American
  Chapter of the Association for Computational Linguistics: Human Language
  Technologies}, pages 515--520.

\bibitem[{Liu and Lane(2016)}]{liu}
Bing Liu and Ian Lane. 2016.
\newblock Attention-based recurrent neural network models for joint intent
  detection and slot filling.
\newblock \emph{Proceedings of The 17th Annual Meeting of the International
  Speech Communication Association (INTERSPEECH 2016)}.

\bibitem[{Liu et~al.(2019{\natexlab{a}})Liu, He, Chen, and Gao}]{mtdnn}
Xiaodong Liu, Pengcheng He, Weizhu Chen, and Jianfeng Gao. 2019{\natexlab{a}}.
\newblock \href {http://arxiv.org/abs/1901.11504} {Multi-task deep neural
  networks for natural language understanding}.
\newblock \emph{CoRR}, abs/1901.11504.

\bibitem[{Liu et~al.(2019{\natexlab{b}})Liu, Ott, Goyal, Du, Joshi, Chen, Levy,
  Lewis, Zettlemoyer, and Stoyanov}]{roberta}
Yinhan Liu, Myle Ott, Naman Goyal, Jingfei Du, Mandar Joshi, Danqi Chen, Omer
  Levy, Mike Lewis, Luke Zettlemoyer, and Veselin Stoyanov. 2019{\natexlab{b}}.
\newblock \href {http://arxiv.org/abs/1907.11692} {Roberta: {A} robustly
  optimized {BERT} pretraining approach}.
\newblock \emph{CoRR}, abs/1907.11692.

\bibitem[{McCarley(2019)}]{prune_bert}
J.~Scott McCarley. 2019.
\newblock Pruning a bert-based question answering model.
\newblock \emph{ArXiv}, abs/1910.06360.

\bibitem[{Ortega and Vu(2017)}]{ortega-vu:2017:W17-55}
Daniel Ortega and Ngoc~Thang Vu. 2017.
\newblock Neural-based context representation learning for dialog act
  classification.
\newblock In \emph{Proceedings of the 18th Annual SIGdial Meeting on Discourse
  and Dialogue}, pages 247--252.

\bibitem[{Radford et~al.(2019)Radford, Wu, Child, Luan, Amodei, and
  Sutskever}]{gpt2}
Alec Radford, Jeffrey Wu, Rewon Child, David Luan, Dario Amodei, and Ilya
  Sutskever. 2019.
\newblock Language models are unsupervised multitask learners.

\bibitem[{Ravi(2017)}]{DBLP:journals/corr/abs-1708-00630}
Sujith Ravi. 2017.
\newblock \href {http://arxiv.org/abs/1708.00630} {Projectionnet: Learning
  efficient on-device deep networks using neural projections}.
\newblock \emph{CoRR}, abs/1708.00630.

\bibitem[{Ravi(2019)}]{pmlr-v97-ravi19a}
Sujith Ravi. 2019.
\newblock \href {http://proceedings.mlr.press/v97/ravi19a/ravi19a.pdf}
  {Efficient on-device models using neural projections}.
\newblock In \emph{Proceedings of the 36th International Conference on Machine
  Learning}, volume~97 of \emph{Proceedings of Machine Learning Research},
  pages 5370--5379.

\bibitem[{Ravi and Kozareva(2018)}]{sgnn}
Sujith Ravi and Zornitsa Kozareva. 2018.
\newblock \href {https://www.aclweb.org/anthology/D18-1105} {Self-governing
  neural networks for on-device short text classification}.
\newblock In \emph{Proceedings of the 2018 Conference on Empirical Methods in
  Natural Language Processing, Brussels, Belgium, October 31 - November 4,
  2018}, pages 804--810.

\bibitem[{Ravi and Kozareva(2019)}]{ravi-kozareva-2019-device}
Sujith Ravi and Zornitsa Kozareva. 2019.
\newblock \href {https://doi.org/10.18653/v1/P19-1368} {On-device structured
  and context partitioned projection networks}.
\newblock In \emph{Proceedings of the 57th Annual Meeting of the Association
  for Computational Linguistics}, pages 3784--3793, Florence, Italy.
  Association for Computational Linguistics.

\bibitem[{Sanh et~al.(2019)Sanh, Debut, Chaumond, and Wolf}]{distill_bert}
Victor Sanh, Lysandre Debut, Julien Chaumond, and Thomas Wolf. 2019.
\newblock Distilbert, a distilled version of bert: smaller, faster, cheaper and
  lighter.
\newblock \emph{ArXiv}, abs/1910.01108.

\bibitem[{Sankar et~al.(2019)Sankar, Ravi, and
  Kozareva}]{sankar-etal-2019-transferable}
Chinnadhurai Sankar, Sujith Ravi, and Zornitsa Kozareva. 2019.
\newblock \href {https://doi.org/10.18653/v1/N19-1339} {Transferable neural
  projection representations}.
\newblock In \emph{Proceedings of the 2019 Conference of the North {A}merican
  Chapter of the Association for Computational Linguistics: Human Language
  Technologies, Volume 1 (Long and Short Papers)}, pages 3355--3360,
  Minneapolis, Minnesota. Association for Computational Linguistics.

\bibitem[{Shriberg et~al.(2004)Shriberg, Dhillon, Bhagat, Ang, and
  Carvey}]{MRDA}
Elizabeth Shriberg, Rajdip Dhillon, Sonali Bhagat, Jeremy Ang, and Hannah
  Carvey. 2004.
\newblock The {ICSI} meeting recorder dialog act {(MRDA)} corpus.
\newblock In \emph{Proceedings of the {SIGDIAL} 2004 Workshop, The 5th Annual
  Meeting of the Special Interest Group on Discourse and Dialogue, April 30 -
  May 1, 2004, Cambridge, Massachusetts, {USA}}, pages 97--100.

\bibitem[{T{\"u}r et~al.(2010)T{\"u}r, Hakkani-T{\"u}r, and Heck}]{atis}
G{\"o}khan T{\"u}r, Dilek Hakkani-T{\"u}r, and Larry~P. Heck. 2010.
\newblock What is left to be understood in atis?
\newblock In \emph{Proceedings of 2010 IEEE Spoken Language Technology Workshop
  (SLT)}, pages 19--24.

\bibitem[{Vaswani et~al.(2017)Vaswani, Shazeer, Parmar, Uszkoreit, Jones,
  Gomez, Kaiser, and Polosukhin}]{vaswani2017attention}
Ashish Vaswani, Noam Shazeer, Niki Parmar, Jakob Uszkoreit, Llion Jones,
  Aidan~N Gomez, {\L}ukasz Kaiser, and Illia Polosukhin. 2017.
\newblock Attention is all you need.
\newblock In \emph{Advances in Neural Information Processing Systems}, pages
  5998--6008.

\bibitem[{Wang et~al.(2018)Wang, Singh, Michael, Hill, Levy, and Bowman}]{glue}
Alex Wang, Amanpreet Singh, Julian Michael, Felix Hill, Omer Levy, and
  Samuel~R. Bowman. 2018.
\newblock \href {http://arxiv.org/abs/1804.07461} {{GLUE:} {A} multi-task
  benchmark and analysis platform for natural language understanding}.
\newblock \emph{CoRR}, abs/1804.07461.

\bibitem[{Yang et~al.(2019)Yang, Dai, Yang, Carbonell, Salakhutdinov, and
  Le}]{xlnet}
Zhilin Yang, Zihang Dai, Yiming Yang, Jaime~G. Carbonell, Ruslan Salakhutdinov,
  and Quoc~V. Le. 2019.
\newblock \href {http://arxiv.org/abs/1906.08237} {Xlnet: Generalized
  autoregressive pretraining for language understanding}.
\newblock \emph{CoRR}, abs/1906.08237.

\bibitem[{Zhang et~al.(2015{\natexlab{a}})Zhang, Zhao, and
  LeCun}]{zhang2015text}
Xiang Zhang, Junbo Zhao, and Yann LeCun. 2015{\natexlab{a}}.
\newblock Character-level convolutional networks for text classification.
\newblock In \emph{Advances in Neural Information Processing Systems}, pages
  649--657.

\bibitem[{Zhang et~al.(2015{\natexlab{b}})Zhang, Zhao, and LeCun}]{Zhang:2015}
Xiang Zhang, Junbo Zhao, and Yann LeCun. 2015{\natexlab{b}}.
\newblock Character-level convolutional networks for text classification.
\newblock In \emph{Proceedings of the 28th International Conference on Neural
  Information Processing Systems - Volume 1}, NIPS'15, pages 649--657. MIT
  Press.

\bibitem[{Zhu et~al.(2015)Zhu, Kiros, Zemel, Salakhutdinov, Urtasun, Torralba,
  and Fidler}]{bookcorpus}
Yukun Zhu, Ryan Kiros, Richard~S. Zemel, Ruslan Salakhutdinov, Raquel Urtasun,
  Antonio Torralba, and Sanja Fidler. 2015.
\newblock \href {https://doi.org/10.1109/ICCV.2015.11} {Aligning books and
  movies: Towards story-like visual explanations by watching movies and reading
  books}.
\newblock In \emph{2015 {IEEE} International Conference on Computer Vision,
  {ICCV} 2015, Santiago, Chile, December 7-13, 2015}, pages 19--27.

\end{thebibliography}
\bibliographystyle{acl_natbib}

\appendix

\end{document}